\documentclass[10pt,twocolumn,letterpaper]{article}

\usepackage{cvpr}              

\usepackage[utf8]{inputenc} 
\usepackage[T1]{fontenc}    
\usepackage[pagebackref,breaklinks,colorlinks]{hyperref}
\usepackage{url}            
\usepackage{booktabs}       
\usepackage{amsfonts}       
\usepackage{amsmath}
\usepackage{amssymb}
\usepackage{graphicx}
\usepackage{nicefrac}       
\usepackage{microtype}      
\usepackage{subfiles}
\usepackage{dsfont}
\usepackage{multirow}
\usepackage{xcolor}

\makeatletter
\@namedef{ver@everyshi.sty}{}
\makeatother
\usepackage{tikz}
\usepackage{tikz-network}

\usepackage[capitalize]{cleveref}
\crefname{section}{Sec.}{Secs.}
\Crefname{section}{Section}{Sections}
\Crefname{table}{Table}{Tables}
\crefname{table}{Tab.}{Tabs.}

\usepackage{subfiles}

\newcommand{\bx}{\mathbf{x}}
\newcommand{\bz}{\mathbf{z}}

\def\cvprPaperID{57} 
\def\confName{CVPR}
\def\confYear{2022}

\begin{document}

\title{Self-Supervised Pre-training of Vision Transformers for Dense Prediction Tasks}

\author{Jaonary Rabarisoa \hspace{2em} Valentin Belissen \hspace{2em} Florian Chabot \hspace{2em}  Quoc-Cuong Pham\\
Universit\'e Paris-Saclay, CEA, List, F-91120, Palaiseau, France\\
{\tt\small \{firstname.lastname\}@cea.fr}
}
\maketitle

\begin{abstract}
    We present a new self-supervised pre-training of Vision Transformers for dense prediction tasks. It is based on a contrastive loss across views that compares pixel-level representations to global image representations. This strategy produces better local features suitable for dense prediction tasks as opposed to contrastive pre-training based on global image representation only. Furthermore, our approach does not suffer from a reduced batch size since the number of negative examples needed in the contrastive loss is in the order of the number of local features. We demonstrate the effectiveness of our pre-training strategy on two dense prediction tasks: semantic segmentation and monocular depth estimation.
\end{abstract}


\section{Introduction}
\label{sec:introduction}

Recently, Vision Transformers \cite{dosovitskiy2021image} (ViT) have become a powerful alternative to Convolutional Neural Networks (CNN) when solving computer vision tasks. On several benchmarks (image classification, object detection, semantic segmentation, monocular depth prediction, ...) most of the top performing methods use a ViT as a core component. This performance can be explained by a reduced inductive bias in ViT, compared to CNN, which leads to better generalization capability. But this comes at a cost: the need for a large annotated dataset \cite{steiner2021train}. For a given task, when the amount of training data is limited, pre-training combined with transfer learning is the most successful way to use ViTs.

Several works deal with pre-training ViT models. They can be supervised \cite{steiner2021train} by solving a classification task on a large dataset \cite{Russ2014Imagenet,ridnik2021imagenet21k} or self-supervised \cite{touvron2021training, caron2021emerging, chen2021empirical}. Generally, these approaches seek to learn a global representation at the image level during the pre-training phase, which is then used as initialization to different downstream tasks. Even though they exhibit a good performance when transferred to a pixel-level task such as semantic segmentation \cite{grill2020bootstrap}, we argue that their performance could be improved by taking into account the local aspect of the downstream task.

In this work, we propose to learn discriminative local features using a new self-supervised learning approach. Our pretext task compares representations of every part of a signal to its global representation using contrastive loss \cite{oord2019representation,chen2020simple}. With this strategy, we learn part representations that are informative about the global context allowing a better initialization for future dense downstream tasks. For instance, in semantic segmentation we predict the class of every pixel of an image. The classes to be predicted have high semantic level and define a global concept on the image. The representation of each pixel should then have a sufficient information about the class it belongs to. Several lines of work have demonstrated that contrasting local and global views is a good strategy to learn a representation. In \cite{hassani2020contrastgraph} the authors use it to learn nodes and a graph representation. The multi-crop strategy presented in \cite{caron2020swav} compares several views generated at different scale levels. Another benefit in contrasting local and global views is that we naturally have several local views and then have access to more negatives samples. Our approach is then less sensitive to the batch size. Finally, as shown in \cite{Khosla2020supcon}, using several positives in the contrastive loss \cite{chen2020simple} helps to learn a better representation.

Our contribution is two-fold: 1) We propose a new pre-training strategy of Vision Transformers \cite{dosovitskiy2021image} designed for dense prediction tasks based on local to global contrastive learning. 2) We present a comprehensive study to demonstrate the effectiveness of our pre-training method on the tasks of semantic segmentation and monocular depth estimation.

\section{Related Work}
\label{sec:related_work}

\paragraph{Self-supervised learning} Self-supervised learning is a set of techniques used to pre-train a neural network with unlabelled data before solving a specific downstream task. The most successful approaches use contrastive representation learning \cite{oord2019representation}. They predict the next representation of a sequence from the current context by solving a classification problem where the target representation is the positive class and any other vectors from the dataset are considered as negative. A standard loss function is the Noise-Contrastive Estimation loss (InfoNCE)  \cite{gutmann10a} which is shown to be a lower bound of the mutual information between the current context and the next features in the sequence. Based on this loss, SimCLR \cite{chen2020simple,chen2020simclr2} learns visual representations by contrasting different views of the same image with other images, by using carefully designed random augmentations and a large training batch size. The method outperforms supervised pre-training in image classification tasks. MoCo \cite{He2020moco} alleviates the need of a large batch size by sampling the negative examples from a memory bank and uses a momentum network to compute the target feature vectors. AMDIM \cite{bachman2019learning} is a multi-scale extension of contrastive representation learning from multiple views. It predicts local representations of a CNN feature map from global description vectors at different levels of the network. Having a large number of good quality negative examples is a key factor in the convergence of InfoNCE loss optimization. Non-contrastive approaches propose to solve the multi-view representation learning problem without negative samples. BYOL \cite{grill2020bootstrap} directly matches the outputs of a Siamese network using mean squared error. SwAV \cite{caron2020swav} learns the representation by predicting a pseudo-label from one view using the other view. Pseudo-labels are computed by clustering representations with Sinkhorn's algorithm. Barlow Twins \cite{zbontar2021barlow} minimizes an objective function based on the cross-correlation matrix of the two views.

\paragraph{Pre-training Vision Transformers} The performance of ViT in computer vision tasks highly depends on the pre-training phase, especially when the training data is small. Steiner \textit{et al.} in \cite{steiner2021train} study the inter-play between regularization, data augmentation and dataset size when training ViTs for image classification. DeiT \cite{touvron2021training} is a data-efficient training method using distillation through attention. It shows competitive results when training ViT models on ImageNet-1k with no external data.
Other works study the self-supervised pre-training of ViT. DINO \cite{caron2021emerging} belongs to the multi-view representation learning methods. It uses self-distillation and pseudo-labels to learn a visual representation. The authors highlight the importance of local-to-global correspondence through the use of the multi-crop strategy \cite{caron2020swav}. In BeIT \cite{bao2021beit}, inspired by BERT \cite{devlin2019bert}, image patches are encoded into discrete visual tokens and the pre-training objective is to recover these visual tokens from corrupted patch embeddings.

\paragraph{Transformers for Dense Prediction} DETR \cite{carion2020detr} is one of the first successful applications of Transformers for dense prediction tasks. It uses a transformer encoder-decoder network to directly predict object location and class without any extra components required in standard object detection pipelines. Segmenter \cite{strudel2021segmenter} uses ViT as its encoder network and a transformer decoder network similar to the one used in DETR in order to predict semantic segmentation masks. Although we follow a similar architecture design as Segmenter, we mainly focus on the pre-training of the encoder. 

Applying Transformers to dense vision tasks raises computational complexity issues especially for high-resolution images, as self-attention has quadratic complexity with respect to the number of input tokens. More efficient architectures have been introduced (Swin Tranformer \cite{liu2021swin}, Focal Transformer \cite{yang2021focal}, CSWin Transformer \cite{dong2021cswin}, DPT \cite{ranftl2021vision}) and are now state-of-the-art for dense prediction tasks. In all of these works, the pre-training phase and the use of large datasets remain critical to reach optimal performance.

In our pre-training approach, we generalize the InfoNCE loss with several positive examples \cite{Khosla2020supcon} to learn to predict local features from a global image representation. As opposed to AMDIM, our method only works with the external features outputted by the ViT network, which makes it more computationally efficient. Following the same intuition that global-to-local feature prediction  \cite{caron2020swav} allows to learn stronger representations, our approach is not data-driven but only relies on the image patches imposed by the ViT architecture.

\section{Approach}
\label{sec:approach}

\begin{figure}
    \centering
    \begin{tikzpicture}[thick, yscale=0.55, xscale=0.8, every node/.style={scale=0.75}]
       
        \Vertex[x=2, y=-3.5, size=0, opacity=0, style={color=white}]{IM1}
        \node (im1) at (2, -4.70) {\includegraphics[scale=0.08]{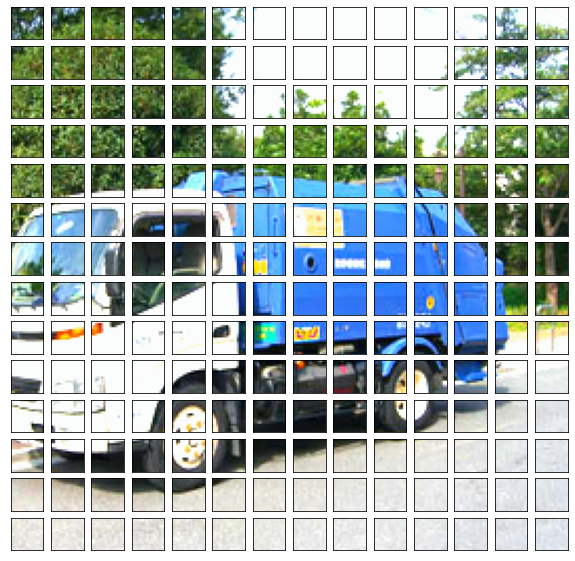}};

        \Vertex[x=2, y=-2.5, size=0, opacity=0, style={color=white}]{E1}
        \draw[fill=orange!75, draw=none] (0.5, -1.5) rectangle node(Embedding) {Patch Embedding} (3.5, -2.5);

        \Edge[Direct, opacity=0.7, lw=0.75](IM1)(E1)

        \Vertex[x=1, y=-1.5, size=0, style={draw=none, color=white, opacity=0}]{POUTB}
        \Vertex[x=3, y=-1.5, size=0, style={draw=none, color=white, opacity=0}]{POUTC}

        \Vertex[y=-0.25, label=cls, style={draw=none, color=blue!50}]{A}
        \Vertex[x=1, y=-0.25, style={draw=none}, label=$\mathbf{x}_1$]{B}
        \Vertex[x=3, y=-0.25, style={draw=none}, label=$\mathbf{x}_N$]{C}

        \Edge[Direct, opacity=0.7, lw=0.75](POUTB)(B)
        \Edge[Direct, opacity=0.7, lw=0.75](POUTC)(C)

        \Vertex[y=1, size=0, opacity=0, style={color=white, opacity=0}]{AIN}
        \Vertex[x=1, y=1, size=0, style={color=white, opacity=0}]{BIN}
        \Vertex[x=3, y=1, size=0, style={color=white, opacity=0}]{CIN}

        \Edge[Direct, opacity=0.7, lw=0.75](A)(AIN)
        \Edge[Direct, opacity=0.7, lw=0.75](B)(BIN)
        \Edge[Direct, opacity=0.7, lw=0.75](C)(CIN)
        
        \Vertex[y=3.25, style={draw=none, color=blue!50}, label=$\mathbf{z}$]{AA}
        \Vertex[x=1,y=3.25, style={draw=none}, label=$\mathbf{z}_1$]{BB}
        \Vertex[x=3,y=3.25, style={draw=none}, label=$\mathbf{z}_N$]{CC}
        
        \Vertex[y=2, size=0,      style={color=white, opacity=0}]{AOUT}
        \Vertex[x=1, y=2, size=0, style={color=white, opacity=0}]{BOUT}
        \Vertex[x=3, y=2, size=0, style={color=white, opacity=0}]{COUT}
    
        \Edge[Direct, opacity=0.7, lw=0.75](AOUT)(AA)
        \Edge[Direct, opacity=0.7, lw=0.75](BOUT)(BB)
        \Edge[Direct, opacity=0.7, lw=0.75](COUT)(CC)

        \draw[fill=orange!75, draw=none] (-0.5, 2) rectangle node(VIT) {Vision Transformer} (3.5, 1);
       
        \Vertex[x=7, y=-3.5, size=0, opacity=0, style={color=white}]{IM2}
        \node (im2) at (7, -4.70) {\includegraphics[scale=0.08]{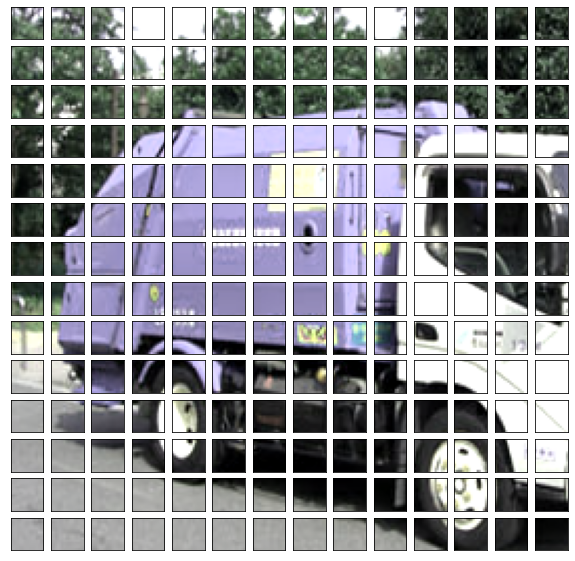}};

        \Vertex[x=7, y=-2.5, size=0, opacity=0, style={color=white}]{E2}
        \draw[fill=orange!75, draw=none] (5.5, -1.5) rectangle node(Embedding) {Patch Embedding} (8.5, -2.5);

        \Edge[Direct, opacity=0.7, lw=0.75](IM2)(E2)

        \Vertex[x=6, y=-1.5, size=0, style={draw=none, color=white, opacity=0}]{POUTB1}
        \Vertex[x=8, y=-1.5, size=0, style={draw=none, color=white, opacity=0}]{POUTC1}

        \Vertex[x=5, y=-0.25, label=$cls^+$
        , style={draw=none, color=blue!50}]{A1}
        \Vertex[x=6, y=-0.25, style={draw=none}, label=$\mathbf{x}_1^+$]{B1}
        \Vertex[x=8, y=-0.25, style={draw=none}, label=$\mathbf{x}_N^+$]{C1}

        \Edge[Direct, opacity=0.7, lw=0.75](POUTB1)(B1)
        \Edge[Direct, opacity=0.7, lw=0.75](POUTC1)(C1)

        \Vertex[x=5, y=1, size=0, opacity=0, style={color=white, opacity=0}]{AIN1}
        \Vertex[x=6, y=1, size=0, style={color=white, opacity=0}]{BIN1}
        \Vertex[x=8, y=1, size=0, style={color=white, opacity=0}]{CIN1}

        \Edge[Direct, opacity=0.7, lw=0.75](A1)(AIN1)
        \Edge[Direct, opacity=0.7, lw=0.75](B1)(BIN1)
        \Edge[Direct, opacity=0.7, lw=0.75](C1)(CIN1)
        
        \Vertex[x=5, y=3.25, style={draw=none, color=blue!50}, label=$\mathbf{z}^+$]{AA1}
        \Vertex[x=6,y=3.25, style={draw=none}, label=$\mathbf{z}_1^+$]{BB1}
        \Vertex[x=8,y=3.25, style={draw=none}, label=$\mathbf{z}_N^+$]{CC1}
        
        \Vertex[x=5, y=2, size=0,      style={color=white, opacity=0}]{AOUT1}
        \Vertex[x=6, y=2, size=0, style={color=white, opacity=0}]{BOUT1}
        \Vertex[x=8, y=2, size=0, style={color=white, opacity=0}]{COUT1}
    
        \Edge[Direct, opacity=0.7, lw=0.75](AOUT1)(AA1)
        \Edge[Direct, opacity=0.7, lw=0.75](BOUT1)(BB1)
        \Edge[Direct, opacity=0.7, lw=0.75](COUT1)(CC1)

        \draw[fill=orange!75, draw=none] (4.5, 2) rectangle node(VIT) {Vision Transformer} (8.5, 1);
       
        \Vertex[x=4, y=5.5, shape=rectangle, style={draw=none, color=green}, label=$\mathcal{L}$]{L}
        \Edge[Direct, opacity=0.7, lw=0.75, bend=30](AA)(L)
        \Edge[Direct, opacity=0.7, lw=0.75, bend=-30](BB1)(L)
        \Edge[Direct, opacity=0.7, lw=0.75, bend=-30](CC1)(L)

    \end{tikzpicture}
    \caption{Two augmented views of the same image are generated using random augmentation. Patches of size $16\times 16$ are extracted from each view and encoded into visual tokens. A learnable class token is used for the global representation. To each token we add its positional encoding and the results are encoded with a ViT. Our loss function contrasts the global representation $\bz$ with all patches representation $\bz_i^+$.}
    \label{fig:overview}
\end{figure}
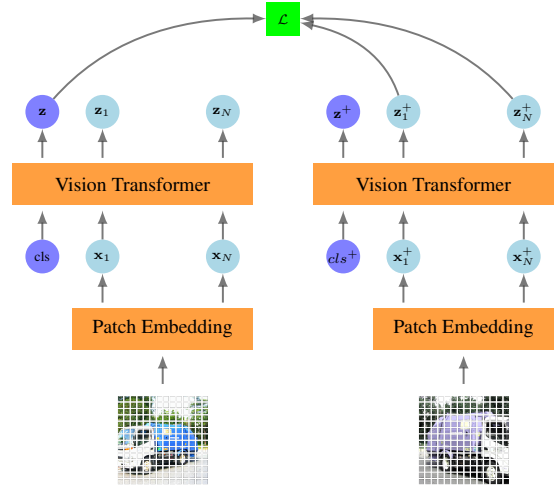

Our goal is to learn a pixel-level representation suitable for dense vision prediction tasks in a self-supervised way. In the following, we denote $\bx = \{\bx_i\}_{i=1\dots N}$ an image with $N$ the number of pixels. Each pixel $\bx_i$ belongs to the RGB space. Let $\bx^+ = \mathcal{T}(\bx)$ be an another view of $\bx$ obtained by randomly transforming $\bx$ with $\mathcal{T}$. The latter is sampled from a set of admissible transformations of $\bx$. 

\subsection{Vanilla Constrastive Learning}
The vanilla contrastive learning learns a global feature transform $\Phi$ that maps the image $\bx$ into a feature vector $\bz = \Phi(\bx) \in \mathbb{R}^d$ by minimizing the InfoNCE objective: 
\begin{equation}
    \label{eq:clr}
    \mathcal{L}_{nce}(\bz, \bz^+) = - \log \frac{e^{\bz^t \bz^+/\tau}}{e^{\bz^t \bz^+/\tau} + \sum_{ \bz^- \in \mathcal{N}(\bz)} e^{\bz^t\bz^- / \tau}}
\end{equation}
where $(\bz, \bz^+)$ are the global feature vectors of the two views of $\bx$ and $\mathcal{N}(\bz)$ is the set of negative features. That is the set of the features of any other images except $\bx$ and its different views. $\tau$ is the temperature. Minimizing (\ref{eq:clr}) aims to group the pair of positive samples $(\bz, \bz^+)$ together while pushing all negative features away from $\mathcal{N}(\bz)$. 

In practice, $\Phi$ is a deep neural network and the objective function is minimized using stochastic gradient descent. The negative features for each example are sampled from the mini-batch. Having a large number of negative samples is critical while minimizing the InfoNCE loss. Hence, a large batch size is required during the optimization.

\subsection{Dense Contrastive Learning}
We propose to learn a pixel-level feature-transform $\Psi$ that yields features, $\bz_i = \Psi(\bx_i)$ for each pixel $\bx_i$. To learn $\Psi$ we use the InfoNCE loss (\ref{eq:clr}) that compares local and global features in the following way:

\begin{equation}
    \label{eq:denseclr}
    \mathcal{L} = \frac{1}{N} \sum_{i=1}^N \mathcal{L}_{nce}(\bz, \bz_i^+)
\end{equation}
In equation (\ref{eq:denseclr}), $\bz$ represents the global representation of the image $\bx$ and $\bz_i^+$ a local feature from its corresponding view $\bx^+$. Minimizing (\ref{eq:denseclr}) aims to contrast all the patches representation from one view with the global representation of another view. This pretext task is closely related to the dense prediction task of semantic segmentation where we predict the class of each pixel in a given image. We argue that this helps to learn more meaningful local features since it will contain global information. 

A similar approach has been proposed to learn nodes and a graph level representation in \cite{hassani2020contrastgraph}. While they learn a representation for general graphs we apply it to ViT viewed as a fully connected graph of patches. Also, the multi-crop strategy proposed by \cite{caron2020swav} brings evidence that contrasting local and global views helps to learn robust features. In our case, we do not generate explicitly the local and global views by data augmentation but rather use the structure imposed by the Vision Transformer.

In practice, the optimization is done by stochastic gradient descent and the negative features are sampled from the mini-batch. With this formulation, the number of negative samples used in the InfoNCE is multiplied by the number of patches. This makes our approach less dependent on the batch size.

\section{Experiments}
\label{sec:experiments}

In this section we present our experimental setup and main results. We follow the standard evaluation protocols for self-supervised learning. After the pre-training phase, the learned features are evaluated on two downstream tasks: semantic segmentation and monocular depth estimation. At the evaluation time, we add a task related head on top of the ViTs encoder. The whole network is fine-tuned on the target dataset.

\subsection{Pre-Training}

We study three different Vision Transformer configurations \cite{dosovitskiy2021image,touvron2021training}: ViT-Ti, ViT-S, ViT-B. We set the ViT patch size to $16 \times 16$. The projection network is a 3-layer MLP \cite{chen2020simple} with a \textit{gelu} activation. The temperature of the InfoNCE loss is fixed at $\tau=0.1$. The models are pre-trained with an input resolution of $224\times 224$.

We use AdamW \cite{loshchilov2018decoupled}  as optimizer, a cosine decay learning rate scheduler and a linear warm-up for $5\%$ of the total epochs. The base learning rate is $10^{-4}$ and is linearly scaled with respect to the batch size. In our current implementation, we do not gather the negative examples across all accelerator devices and only use per device negative to compute the InfoNCE loss (\ref{eq:clr}). The batch size per GPU is set to $128$. The pre-training is distributed across 16 Nvidia A100-80G.

By default, we use the same data augmentation policy as SimCLR \cite{chen2020simple}. First, the image is randomly cropped and resized to $224 \times 224$. Then, the image color is randomly distorted and optionally converted to grayscale. Finally, Gaussian blur is randomly applied.

\subsection{Semantic segmentation}

Semantic segmentation aims to predict the class of each pixel in a given image. We conduct our experiments on the ADE20k dataset \cite{zhou2019semantic}. After the pre-training, we append a segmentation head to the pre-trained encoder. The whole network is fine-tuned on the training set. We report results on the validation set. We consider two types of segmentation heads: linear and the UPerNet \cite{xiao2018unified,bao2021beit}. The input resolution is set to $512\times 512$. We use bicubic interpolation to interpolate the positional embedding of the ViT. 

The segmentation model is learned using the pixel-wise cross-entropy objective. We use AdamW as optimizer and a base learning rate of $0.0001$ with a polynomial decay policy. During the $64$ epochs of the fine-tuning, we set the weights decay to $0.005$, the stochastic depth to $0.1$ and we apply a layer-wise decay to the learning rate. 

Table~\ref{res:ade20k} presents our results for different pre-training methods on the ImageNet-1k and several combinations of Encoder/Head networks. For the Supervised and DINO pre-training we use the encoder weights provided by \cite{steiner2021train} and \cite{caron2021emerging} and run the fine-tuning task ourselves. It shows that our pre-training strategy gives the best performance especially for ViT-Ti and ViT-S. We find out that the regularization is critical for the fine-tuning phase when the size of the network increases. For the ViT-B architecture we use a higher value of weights decay and stochastic depth compared to ViT-Ti and ViT-S. Comparatively, a ViT-B pre-trained on ImageNet-21k with BeIT \cite{bao2021beit} performs better than ours. This suggests that our ViT-B still suffers from overfitting and a the use of a large dataset can help the fine-tuning.

\begin{table}[h!]
  \centering
  \begin{tabular}{c c c c}
    \toprule
    Encoder                    & Pre-training                      & Head                         & mIoU                       \\
    \midrule
    \multirow{3}{*}[-2pt]{ViT-Ti/16} & Supervised\**                        & Linear                       & 32.4                      \\
    \cmidrule(lr){2-4}
                               & \multirow{2}{*}{Ours}             & Linear                       & 36.2                       \\
                               &                                   & UPerNet                      & \textbf{38.0}                \\
    \cmidrule(lr){1-4}
    \multirow{4}{*}[-3pt]{ViT-S/16}  & Supervised\**                        & Linear                       & 42.1                      \\
                               & DINO\**                              & Linear                       & 38.8                      \\
    \cmidrule(lr){2-4}
                               & \multirow{2}{*}{Ours}             & Linear                       & \textbf{42.8}              \\
                               &                                   & UPerNet                      & \textbf{43.2}              \\
    \cmidrule(lr){1-4}
    \multirow{5}{*}[-7pt]{ViT-B/16}  & Supervised\**                        & Linear                       & 43.1                      \\
                               & DINO\**                              & Linear                       & 43.2                      \\
    \cmidrule(lr){2-4}
                               & \multirow{2}{*}{Ours}             & Linear                       & \textbf{45.1}              \\
                               &                                   & UPerNet                      & \textbf{45.1}                       \\
    \cmidrule(lr){2-4}
                               & \textit{BeIT (i-21k)} & \textit{UPerNet} & \textit{\textbf{53.6}} \\
    \bottomrule
  \end{tabular}
  \caption{Performance of the ViT models pre-trained on ImageNet-1k and fine-tuned on ADE20K. We report the mIoU on the validation set. (\**) Models are fine-tuned from the weights provided by the authors. (i-21k) indicates pre-training done on ImageNet-21k and result from the paper \cite{bao2021beit}.}
  \label{res:ade20k}
\end{table}

\subsection{Monocular depth estimation}

We evaluate the generalization capability of the pre-trained representation by fine-tuning the network on a depth estimation task on the NYU-Depth V2 dataset \cite{silberman2012indoor}. 

Two kinds of regression heads are experimented: on the one hand, a linear series of 4 up-projection convolutions with a stride of 2 until the image resolution is retrieved. On the other hand, we experiment a more sophisticated head inspired from the work of \cite{xiao2018unified} (UPerNet). In both cases, the final activation function is a dilated sigmoid that matches the depth range of the NYU-Depth V2 dataset. The training loss is a berHu loss like in \cite{laina2016deeper} regularized with a depth smoothness term ($\mathds{L}_2$ gradient loss, inspired from \cite{heise2013pm}).

The pre-trained models are fine-tuned on the 47584 image/depth map pairs of the train split, and evaluated on the 654 pairs of the test split. Data augmentation includes random color jitter, crop and horizontal flip. We evaluate our models using standard metrics -- absolute relative error ($\text{AbsRel}$), RMSE, threshold accuracy ($\delta_1$). All models are trained for 50 epochs, with the Adam optimizer and a learning rate of $0.0001$.

\begin{table}[h!]
 \centering
 \begin{tabular}{c c c c c}
 \toprule
\multicolumn{2}{c}{Method} & AbsRel & RMSE & $\delta_1$\\
 \midrule
 \multicolumn{2}{c}{FCRN-Depth \cite{laina2016deeper}} & 0.127 & 0.573 & 0.811 \\
 \multicolumn{2}{c}{SimCLR \cite{chen2020simple}} & 0.134 & 0.557 & 0.833  \\
 \multicolumn{2}{c}{BYOL \cite{grill2020bootstrap}} & 0.129 & 0.541 & 0.846 \\
 \cmidrule(lr){1-5}
\multirow{2}{*}{ViT-Ti/16} & Linear & 0.140 & 0.598 & 0.823 \\
\vspace{2pt}
& UPerNet & 0.138 & 0.593 & 0.832  \\
\multirow{2}{*}{ViT-S/16} & Linear & 0.124 & 0.564 & 0.856  \\
\vspace{2pt}
& UPerNet & \textbf{0.122} & 0.549 & 0.862  \\
\multirow{2}{*}{ViT-B/16} & Linear & 0.123 & 0.544  & 0.862 \\
& UPerNet & \textbf{0.122} & \textbf{0.526} & \textbf{0.865}\\
 \bottomrule
 \end{tabular}
 \caption{Performance of the pre-trained ViT models when fine-tuned on a downstream depth estimation task (NYU-Depth V2 dataset). For the tiny (ViT-Ti), small (ViT-S) and big (ViT-B) versions, both a linear and UPerNet-style heads are evaluated, with better performance than state-of-the-art reference points.}
 \label{tab:depthprediction}
\end{table}

Our results are presented in Table~\ref{tab:depthprediction}. BYOL \cite{grill2020bootstrap} and SimCLR \cite{chen2020simple} are the two main self-supervised reference points. These two models have indeed been pre-trained on the training set of the ImageNet-1k dataset \cite{russakovsky2015imagenet} -- without using labels -- and fine-tuned on the NYU-Depth V2 dataset. As for \cite{grill2020bootstrap} and for a broader perspective, we also include results from FCRN-Depth \cite{laina2016deeper}, which uses supervised pre-training on ImageNet-1k before fine-tuning on NYU-Depth V2. However, since we aim at assessing the performance of the fully self-supervised pre-training and for a fair comparison, we do not report the results of more recent methods that mix ViTs and CNN representations with weights trained in a supervised fashion on large image datasets, like \cite{yang2021transformer}. For the same reason, we also exclude results from models pre-trained on much larger depth datasets, like \cite{ranftl2021vision}.

Our method outperforms the self-supervised state-of-the-art with respect to all metrics, an observation that holds true for almost all metrics even when using the smaller ViT-S/16 or with the linear up-convolution heads.

\section{Conclusion}

We have presented a new method to pre-train Vison Transformers for dense vision tasks. It is an extension of the contrastive learning framework that compares local to global features. We have shown that this pre-training strategy is competitive on semantic segmentation and monocular depth estimation tasks. For future work we plan to study the effect of more advanced data augmentation on this approach and the use of non contrastive losses as objective function.

\section{Acknowledgments}
This work benefited from the FactoryIA supercomputer financially supported by the Ile-deFrance Regional Council.

{\small
\bibliographystyle{plain}
\bibliography{main}
}

\end{document}